# Book: "Diabetes Digital Health, Telehealth, and Artificial Intelligence". Chapter: "Food Recognition and Nutritional Apps"

*Lubnaa Abdur Rahman[1], Ioannis Papathanail[1], Lorenzo Brigato[1], Elias K. Spanakis[2], Stavroula Mougiakakou[1,3]*

[1]ARTORG Center for Biomedical Engineering Research, University of Bern, Bern, Switzerland,

[2] Division of Endocrinology, Diabetes, and Nutrition, University of Maryland School of Medicine, Baltimore, MD, USA,

[3]Department of Emergency Medicine, Inselspital, University Hospital, University of Bern, Bern, Switzerland

## Abstract

Food recognition and nutritional apps are trending technologies that may revolutionise the way people with diabetes manage their diet. Such apps can monitor food intake as a digital diary and even employ artificial intelligence to assess the diet automatically. Although these apps offer a promising solution for managing diabetes, they are rarely used by patients. This chapter aims to provide an in-depth assessment of the current status of apps for food recognition and nutrition, to identify factors that may inhibit or facilitate their use, while it is accompanied by an outline of relevant research and development.

## Abbreviations

**AI**, Artificial Intelligence
**CHO**, Carbohydrates
**HCP,** Healthcare provider
**PwD**, People with Diabetes

**Summary**

- Mobile apps for food recognition and nutrition are based on user input and/or meal images or videos. They provide the nutritional content of a meal and can thus help people with diabetes to monitor their diet. There has been little systematic evaluation of these apps.
- We have performed a quantitative and qualitative assessment of the current mobile apps to assess diet, with realistic scenarios.
- Some of the tested apps can automatically predict the nutrient content of a meal with minimal input from the user, but they have several limitations - such as incorrect food recognition, inaccurate volume estimation, and poor user engagement.

**Statistics**

- People with diabetes tend to underestimate their carbohydrate intake, particularly for meals with high carbohydrate intake, by 28-34% [1]. Another study showed that for 61 participants, the average accuracy of the carb counting was of only 59% [2]
- Studies have shown that, out of 1001 healthcare providers, 45.5% recommend food and nutrition apps to their patients [3]
- Out of the 60 nutritional apps that we went through in our experiment, only 10 provided AI capabilities out of which only four featured an automatic end-to-end pipeline for dietary assessment.

**Introduction**

Diabetes remains a global health concern, with approximately 537 million people affected in 2021 [4]. Managing the condition remains an overwhelming task for "People with Diabetes" (PwD), as patients must constantly monitor numerous health parameters, including glucose levels, food consumption, and physical activity [5]. With the current advances in both Artificial Intelligence (AI) and computer vision, digital health technologies promise to support PwD in managing their condition and easing their burden [6]. One prime approach is to use apps for food recognition and nutrition, as these can automatically assess diet and estimate the carbohydrate (CHO) content of a meal from the user input, meal images or videos [7]. One study demonstrated that PwD underestimate carbohydrate intake by 28-34% while another one pointed out that for 61 participants the carb counting accuracy of PwD was only of 59% [1,2]. While there is an abundance of apps food recognition and nutrition, and even though as studies have shown, dieticians recommend the use of these apps to their patients, they are only rarely used for diabetes management, as there is little public information on their correctness and efficacy [3].

**Current Status**

Several mobile apps for food recognition and nutrition are commercially available to help individuals to manage their diets and to enhance their awareness of their nutrient intake. However, their levels of automation are disparate. Most apps provide food search within a database, with barcode scanning, but some also allow users to take pictures of meals for food recognition and to further assess the nutritional content. An automatic app for assessing diet in PwD should focus on reducing the burden to users - by minimising manual input (e.g., using images or short videos) for food recognition and assessing nutritional content [8]. The ideal app should feature a fully automated pipeline for dietary assessment, as shown in Fig.1, including food segmentation, recognition, volume estimation, and nutrient content calculation based on food composition databases.

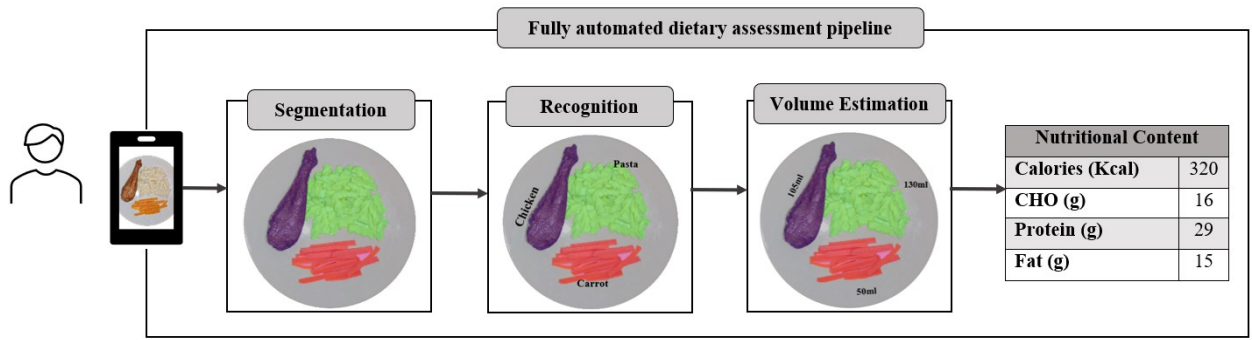

*Figure 1: Fully automated dietary assessment pipeline*

In order to provide an in-depth view of the current market for apps for dietary assessment, and to effectively identify those apps that can be useful for PwD, we first searched online for available nutritional apps, and then conducted an experiment to evaluate their performance in realistic scenarios for estimating content. As background work, we perused popular apps available within Switzerland. Out of the 60 food and nutritional apps found, we filtered out those that could not automatically track or recognise food, leaving us with a total of 10 applications to carry out our experiments with: *Bitesnap*[1] *(ver. 1.7.1), Calorie Mama*[2] (ver. 5.330.41301), MyFitnessPal[3] (ver. 23.7.5.43901), SnackSnap [4] (ver. 2.2.0), Foodvisor[5] (ver. 5.4.0), Undermyfork[6] (ver. 2.24.2), DiabTrend[7] (ver. 2.43.14), DietCameraAI[8] (ver. 2.5.7), goFOOD$^{TM}$ [9] (ver. 1.0), and SNAQ[10] (ver. 8.10.1) [9],[10],[11]. While all the apps feature food recognition, only the last four provide automatic estimation of volume/weight; most of the rest estimate nutritional content for standard portions, so that it may be necessary for users to manually adjust the portions of consumed food. Additionally, it is noteworthy to point out that the *SNAQ* application only features sensor-based food depth estimation for iPhones (models X and above) only. For the experimental evaluation of the apps, on 17.02.2023, we reviewed and evaluated their functionalities and performance on three different standard meals which we bought: 1) hamburger, 2) pasta salad, and 3) chicken curry noodles (images in Table 1). To ensure an objective assessment, we carried out the experiments three times using two different phones (iPhone X and OnePlus 7 Pro), with constant lighting conditions, plating arrangements, and the distance at which the image was taken, unless otherwise required by the apps [12]. We also measured the weight of meals, to provide reliable values for ground truth.

---

[1] https://getbitesnap.com/
[2] https://www.caloriemama.ai/

**Food Recognition**

While testing the food recognition module, we found that all the apps are able to correctly identify the burger, even if some were better than others at identifying the specific type of burger, such as "hamburger double patty". *Calorie Mama* and *DiabTrend* were even able to distinguish the fast-food chain the burger came from, and both predicted that the burger was a McDonald's Big Mac, which is an interesting feature that allows for more precise nutritional content. On the other hand, *SNAQ*, *Bitesnap*, and *DietCameraAI* had the option for users to refine the results for finding the burger corresponding to the fast-food chain, but this required manual intervention like a search from the database. For the pasta salad, we found that the *goFOOD$^{TM}$* presented the most accurate prediction, while the other apps were only able to identify parts of the dish on the plate, such as either pasta or salad, rather than as a whole meal. *SnackSnap* failed to provide a specific prediction, as it only focussed on one item at a time on the plate and, therefore, could only identify one object on the plate and then provided "tomato" as its prediction.

---

[3] https://www.myfitnesspal.com/
[4] https://snacksnap.app/
[5] https://www.foodvisor.io/en/
[6] https://undermyfork.com/
[7] https://diabtrend.com/
[8] https://www.doinglab.com/en/theme/s007/index/company_02.php
[9] https://go-food.tech/
[10] https://snaq.io/

*goFOOD$^{TM}$*, *Calorie Mama*, *SNAQ*, and *Foodvisor* were the only apps that correctly identified the last and most challenging dish (chicken curry noodles), even though the predicted category names did not fully match, since different food databases were used. Some of the apps were only able to identify parts of the dish and suggested coarse-grained categories such as noodles or pasta. *DiabTrend* performed poorly, as it kept changing its predictions in real-time, with slight changes in the camera position, and would only look at certain items on the plate. *Bitesnap* required manual selection of each of the ingredients, which was time-consuming and not straightforward. *MyFitnessPal* suggested some items that were partly correct, but since we wanted the closest option to the ground truth, we had to enter the meal manually, using the database search, rather than selecting it from the prediction list.

All applications, except for *SnackSnap*, allowed users to change the suggested food and select from predefined options - either from possible predictions suggested by the app or through searching the database. It is worth noting that some apps, including *Calorie Mama* and *BiteSnap*, only provided coarse categories for meals, and required users to select fine-grained classes in most cases, in order to achieve a more accurate estimation of the ground truth. *goFOOD$^{TM}$* and *SNAQ* provided an excellent user interface, that displayed food segmentation directly on screen, and thus allows users to understand where the AI behind the app is looking. This is essential for explainability and interpretability, as it helps build trust with the end user [8]. Furthermore, *goFOOD$^{TM}$* allowed users to change the segmentation, giving them the option to accept or change the AI's results.

Most of the apps - apart from *goFOOD$^{TM}$* which performed satisfactorily - struggled to consider the meals (especially for meals 2 and 3) as a whole and rather focussed on individual food items. Even though *Undermyfork* presented only very coarse grain categories, it offered a valuable feature that allowed users to choose multiple tags, which could be helpful when multiple ingredients were present in a dish. Nevertheless, this would still require manual input from the user. Overall, in terms of food recognition, we found that in all cases *goFOOD$^{TM}$* was able to give predictions closest to the ground-truth with all three meals, and, in performance, was followed by *Calorie Mama*, *Foodvisor*, and *SNAQ*.

**Automatic Volume/Weight/Nutrient Estimation**

Out of all the food and nutrition apps available on the market, very few of them featured automatic nutrient estimation. To compare the evaluation of nutrients, we selected the four apps that could automatically estimate weight and CHO in grams (g) and energy in kilocalories (Kcal) and further consider to what extent they can support diabetic people. We note that the apps that are not presented in this section (*Calorie Mama*, *Bitesnap*, *SnackSnap*, *Foodvisor*, and *MyFitnessPal*) all provided nutritional information for standard portion sizes that the user could adjust by hand. The exception is *Undermyfork*, which required fully manual input for the nutrient content.

We present our results, in Table 1, in terms of the mean absolute error (MAE) - the absolute deviation between the ground truth and the estimations, so that a lower MAE indicates more accurate estimation of the ground truth. It is evident that major variations were found, however, in 5 out of the 9 cases, *goFOOD*$^{TM}$ gave the lowest MAE values and is either the best or second best with its estimations. *SNAQ* comes after *goFOOD*$^{TM}$, even though some of its MAE values are quite high, for instance for the energy content of the pasta salad (MAE of 215.9g). *DiabTrend* and *DietCameraAI* gave unsatisfactory results and large errors in some estimations, for example in the estimation of the energy of the third meal. During our experiment we further noted that even for standard portion sizes (e.g., 100g), the nutritional information for the same food items often differed between the apps, even though the food category remained the same.

We acknowledge that further studies on larger scales would be needed to provide additional information on the performance of these applications under varying conditions and with different food items. However, we think that our empirical investigation presents a useful initial glimpse into the capabilities of the four tested mobile applications.

|  | Mean Absolute Error of Predicted Values $$MAE = \frac{\sum |predicted\ weight - ground\ truth\ weight|}{number\ of\ times\ experiment\ was\ repeated}$$ | | | |
|---|---|---|---|---|
| Food Images / Ground Truth | *DiabTrend* | *DietCameraAI* | *goFOOD$^{TM}$* | *SNAQ* |
| **Burger** 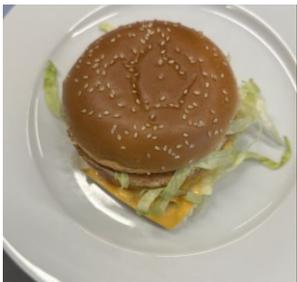 Weight (g): 220 | 87.7 | 128.7 | 23.3 | **21.7** |
| Energy (kcal): 481.8 | 174.8 | 220.1 | **14.8** | 31.1 |
| CHO (g): 39.6 | 11.9 | 17.6 | 4.8 | **2.6** |
| **Pasta Salad** 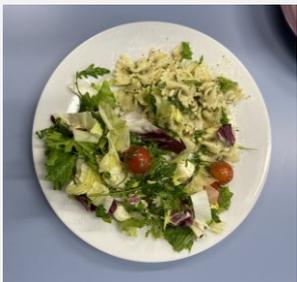 Weight (g): 260 | 92.0 | 53.7 | **32.3** | 132.3 |
| Energy (kcal): 353.6 | **40.1** | 361.7 | 71.9 | 215.9 |
| CHO (g): 20 | 31.3 | 47.0 | **2.4** | 5.0 |
| **Chicken Curry Noodle** 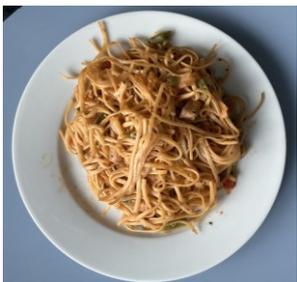 Weight (g): 502 | 181.7 | **34.0** | 36.7 | 229.7 |
| Energy (kcal): 1169.7 | 700.4 | 559.7 | **376.7** | 437.7 |
| CHO (g): 185.7 | 145.0 | 92.7 | **27.1** | 97.4 |

*Table 1: Mean absolute error in automatic estimation of nutrient content by the four apps.*

**Discussion**

In this section, we provide a qualitative evaluation of the tested applications and report the major issues encountered during the experiments, together with the attractive features that the apps presented. Complex foods like salads or layered foods such as noodles mixed with other ingredients were difficult to recognise. Most of the apps permit users to make changes if the results were unsatisfactory. We found that the apps gave different nutritional values for the same foods, presumably because they used different food composition databases. While it was rather easy and straightforward to take the meal pictures using the apps, we observed that slight variations in the angle of the images captured using *DiabTrend* could lead to vastly different weight estimations, some of which differed significantly from the actual values. In the case of *DietCameraAI*, the phone had to be shifted considerably to assure that the portion estimation was properly activated, a feature that was allowed only with Android phones. With *SNAQ,* one of the major limitations was that the feature for automatic portion estimation was restricted to specific iPhone models, and even with the iPhone X, we could only obtain results with the front facing camera, which was not very practical for taking meal pictures. On the other hand, even though *goFOOD$^{TM}$* required the use of two pictures and a reference card while taking pictures, it provided clear instructions on how to position the phone to capture the images correctly - an essential feature absent in the other apps. This could be particularly useful for individuals who are unfamiliar with the specifics of estimating food portion. The time needed from taking the picture to getting the final output was on average 12-15 seconds except for *DietCameraAI* which sometimes took longer since we had to move the phone for it to find the best points.

Most of the applications we evaluated had user-friendly interfaces and, in most cases, it was quite straightforward to understand the output of the nutritional values of the meals. However, this was especially difficult with *Foodvisor*, where multiple questions had to be filled out before it could be used for dietary tracking and assessment. Some readings were difficult to comprehend, as the summary of the nutritional content of a single meal was presented relative to the total consumption required per day. Similarly, with *DietCameraAI*, as its main language used was Korean, it might be hard for some people to comprehend. One major issue in app design generally was that for certain apps, users could only log the meals from the app camera. This means that, if a user had forgotten to log a meal via the app, even if they had a picture of the meal in their gallery, they would still have to add it manually, which would increase the burden on users.

We also note that some features present in certain of the apps could be extremely useful for diabetes patients. One example was the availability of a community feature - as provided by *Calorie Mama*, *MyFitnessPal*, and *SNAQ*. This allows users to interact with each other, receive tips (in the form of suggestions for healthier eating habits and nutrition), as well as simple user education, as in *DiabTrend*, which could help to identify foods with low and high glycaemic index. Some apps also presented extended additional nutritional parameters, such as the levels of fibre and saturated fats. It was interesting to note that *DiabTrend*, *MyFitnessPal*, *SNAQ*, and *Undermyfork* had a strong focus on diabetes management - by providing glucose monitoring features through the integration of sensors.

With regards to pricing, five of the apps reviewed required a subscription for certain features. For example, *SNAQ* allowed only five meal recordings with the free version, while *DiabTrend* required a premium subscription to use the dietary assessment feature. It is also highlighted that all the apps required internet connectivity to function either for authentication or for image processing.

Table 2 summarises our findings in terms of app features and functionalities. While some apps were more holistic than others, we consider that the main criterion for selecting an app must be the accuracy of the results provided.

|  | Meal Input | Multi-label classification | Volume / Weight Estimation | Extended Nutritional Values | Education & Tips | Community | Glucose Monitoring | Free of Charge |
|---|---|---|---|---|---|---|---|---|
| *Bitesnap* | B, I, FS | ✓ | ✗ | ✓ | ✗ | ✗ | ✗ | ✓ |
| *Calorie Mama* | B, I, FS | ✓ | ✗ | ✗ | ✗ | ✓ | ✗ | ✓ |
| *DiabTrend* | B, I, FS, S | ✓ | ✓ | ✓ | ✓ | ✗ | ✓ | ✗ |
| *DietCameraAI* | I | ✓ | ✓ | ✗ | ✗ | ✗ | ✗ | ✓ |
| *Foodvisor* | B, I, FS | ✗ | ✗ | ✓ | ✗ | ✗ | ✗ | ✗ |
| *goFOOD*[TM] | B, I | ✓ | ✓ | ✗ | ✗ | ✗ | ✗ | ✓ |
| *MyFitnessPal* | B, I, FS | ✓ | ✗ | ✓ | ✗ | ✓ | ✓ | ✗ |
| *SnackSnap* | B, I, FS, S | ✗ | ✗ | ✓ | ✗ | ✗ | ✗ | ✗ |
| *SNAQ* | B, I, FS | ✓ | ✓ | ✗ | ✗ | ✗ | ✓ | ✗ |
| *UndermyFork* | I | ✓ | ✗ | ✗ | ✗ | ✗ | ✓ | ✓ |

*Table 2: Summary by app of features and functionalities. For the meal input: B stands for Barcode, I for Image, FS for Food Search, and S for Speech*

**Barriers to progress**

There has been significant progress in the past years in the field of food recognition, and automatic AI-based apps for dietary assessment have proven to be beneficial in clinical contexts [13]. However, their adoption for diabetes management still faces multiple barriers, as there is scepticism about the apps' accuracy at identifying the meals and estimating the nutrient contents [3]. From a regulatory perspective, patients' safety must be prioritized and if a PwD relies heavily on automatic models, insulin adjustment based on erroneous CHO content estimations can even be fatal [14]. Algorithms behind the automatic dietary assessments are not 100% accurate and may struggle to correctly estimate the CHO content of complex meals (with different cooking methods e.g., fried, or baked/numerous ingredients/layered food e.g., salads) or even drinks [15]. This problem is exacerbated by the lack of extensive and varied food image datasets used for training and validating machine learning models. Moreover, the food composition databases are poorly or not standardised which can significantly reduce the accuracy of estimations. None of the current apps features localization, to differentiate between cooking techniques and cuisines where the variability in ingredient quality and portion sizes can lead to significant errors in estimation [7].

Privacy concerns, such as the risk of data breaches, can also hinder the adoption of food recognition and nutritional apps. PwD may be hesitant to use these technologies if they perceive a lack of transparency or control over their personal health data [16]. For the success of any app, user engagement is important. Studies have shown that PwD have reduced usage of the apps after a certain amount of time. Patients are not actively engaged with the app as in most cases they are not included in the app design process [7]. When dealing with a diverse population, biases in app development and datasets can be a hindrance for adoption from PwD. In addition, many apps require the collection of sensitive data, such as keeping track of food trends, blood glucose levels, insulin doses, and other data that could lead to profiling poses to be of great ethical concern [8].

**Needs to advance the field.**

To overcome these barriers and advance within the field, strategies could be implemented to address specific key areas. For datasets of food images, web crawling can be a useful approach for data collection, but there is no assurance that images are relevant or of high quality. It may be desirable to establish protocols and guidelines for standardising data collection, as this could ensure that the collected images datasets are reliable, and this could help in the development of more robust models for machine learning, able to achieve more accurate recognition within apps. Furthermore, standardisation of food composition databases could help to enhance the accuracy of the estimations of nutrient content and reduce variability between apps. In developing the apps, it is important to adopt user-centric designs, while involving the PwD in the process, and to consider the set of tools and framework for Explainable AI [17]. There should be extensive testing and validation of the apps in terms of performance in real-life conditions, while assessing their impact on the health outcomes of PwD. Apps should be as transparent as possible with regards to collection, sharing, and retention of personal data. These would instil more trust among PwD. In addition to technical improvements, there is also a need for greater collaboration and integration with healthcare systems and providers. It would be desirable to invest in training and education programs if we are to increase awareness and understanding of these apps among PwD and their healthcare providers (HCPs). From the app developers' perspective, for the apps to be sustainable and succeed over the long-term, there is the need to establish sustainable business models and to ensure stable revenue streams. This could include providing certain app features only for subscribed users at a premium and integrating and on-boarding healthcare providers within the loop to further promote the use of the apps.

**Future outlook**

There are significant promising areas of research which could help to ensure the seamless adoption of food and nutritional apps and allow them to play an even greater role in diabetes management. Recent studies have also shown that the automatic CHO estimation by PwD, which is further used as input for insulin dosage adjustment, through dietary assessment app may be less prone to errors than manual estimations [18]. As technology continues to evolve, we could possibly see more advanced algorithms and features that can be tailored to improve the support that the user needs or prefers; for instance, personalised nutrition could be an interesting feature that would be of great help for PwD [19]. Furthermore, integration with

HCPs could enable the apps to become seamlessly integrated into overall plans for diabetes management and thus to provide a more holistic and personalised approach to care. App developers could also further target clearance and approval from regulatory bodies, such as the Food and Drug Administration, which would reassure users. Additionally, it would be extremely useful for PwD if sensor data could be integrated into the apps. Possibilities would include continuous glucose monitoring for tracking glucose levels and smart watches for tracking exercise and sleeping patterns [20]. This will, in turn, facilitate the growth of closed-loop systems that can automatically alter insulin dosage on the basis of real time readings. However, if these possibilities are to be implemented, this will require ongoing attention and investment to address the current limitations and challenges associated with the food recognition and nutrition apps.

**Conclusions**

Food recognition and nutrition apps have the potential to revolutionise diabetes management by reducing the burden on PwD. After reviewing several apps, we have presented the current state of the art. We have clearly identified the major drawbacks that were presented in the apps which could impede their adoption for the management of diabetes. Before the full potential of such apps can be realised, several challenges must be addressed, such as the lack of standardisation of food databases and the difficulty in accurately recognising foods and measuring nutrient content. To overcome these challenges and advance the field, we highlight the need for standardisation of databases and the development of large and diverse datasets, as well as seamless integration with HCPs within the loop. Additionally, future research should focus on integrating sensor data, developing personalised nutrition plans, and evaluating the effectiveness and user experience of different apps and devices. With continued research and innovation, the field of food recognition and nutritional apps could significantly improve the outcomes and quality of life of PwD.

**Acknowledgment**
This work was supported in part by the European Commission and the Swiss Confederation - State Secretariat for Education, Research and Innovation (SERI) within the project 101057730 Mobile Artificial Intelligence Solution for Diabetes Adaptive Care (MELISSA).